\documentclass[11pt]{article}

\usepackage{EMNLP2023}

\usepackage{times}
\usepackage{latexsym}
\usepackage[T1]{fontenc}
\usepackage[utf8]{inputenc}
\usepackage{microtype}
\usepackage{inconsolata}
\usepackage{hyperref}
\usepackage{enumitem}
\usepackage{amsmath, amssymb, amsfonts}
\usepackage{booktabs}
\usepackage{comment}
\usepackage{csquotes}
\usepackage{graphicx}
\usepackage{multirow}
\usepackage{subfig}
\usepackage{xfrac}
\usepackage[dvipsnames]{xcolor}
\usepackage[linesnumbered,ruled,vlined]{algorithm2e}

\SetCommentSty{mycommfont}

\newcommand{\gptsmall}{GPT2-Small}
\newcommand{\gptlarge}{GPT2-Large}

\newcommand{\easydata}{\emph{Hand}}
\newcommand{\harddata}{\emph{2WMH}}

\newlength{\encerr}

\newcommand{\encodingerror}{%
  \hbox{%
    $\blacklozenge$%
    \settowidth{\encerr}{$\blacklozenge$}%
    \hspace{-0.77\encerr}%
    {\color{white}\textsmaller[2]{?}}%
   }%
  \xspace%
 }

 \newcommand\blfootnote[1]{%
  \begingroup
  \renewcommand\thefootnote{}\footnote{#1}%
  \addtocounter{footnote}{-1}%
  \endgroup
}

\DeclareMathOperator*{\softmax}{softmax}

\title{
    Memory Injections: Correcting Multi-Hop Reasoning Failures during Inference in Transformer-Based Language Models
}

\author{%
    Mansi Sakarvadia\textsuperscript{\textnormal{1,*}},
    Aswathy Ajith\textsuperscript{\textnormal{1}}, 
    Arham Khan\textsuperscript{\textnormal{1}}, 
    Daniel Grzenda\textsuperscript{\textnormal{1}},
    \\ 
    {\bf Nathaniel Hudson}\textsuperscript{1,2}, 
    {\bf Andr\'{e} Bauer}\textsuperscript{1,2}, 
    {\bf Kyle Chard}\textsuperscript{1,2}, 
    {\bf Ian Foster}\textsuperscript{1,2} 
    \\ \\
     \textsuperscript{1}University of Chicago,
     \textsuperscript{2}Argonne National Laboratory\\
}

\begin{document}
\maketitle

\begin{abstract}
    Answering multi-hop reasoning questions requires retrieving and synthesizing information from diverse sources. 
    Large Language Models (LLMs) struggle to perform such reasoning consistently. Here we propose an approach to pinpoint and rectify multi-hop reasoning failures through targeted \emph{memory injections} on LLM attention heads. First, we analyze the per-layer activations of GPT-2 models in response to single and multi-hop prompts. We then propose a mechanism that allows users to inject pertinent prompt-specific information, which we refer to as \enquote{memories,} at critical LLM locations during inference. By thus enabling the LLM to incorporate additional relevant information during inference, we enhance the quality of multi-hop prompt completions. 
    We show empirically that a simple, efficient, and targeted memory injection into a key attention layer can often increase the probability of the desired next token in multi-hop tasks, by up to 424\%.\blfootnote{* Correspondence to \href{mailto:sakarvadia@uchicago.edu}{\texttt{sakarvadia@uchicago.edu}}}
    
    %Our work advances the field of interpretability and also opens avenues for further research in knowledge retrieval and editing. By reverse engineering the inner workings of LLMs and enabling targeted corrections, we enable future work on enhancing the transparency and effectiveness of these models for complex reasoning tasks.

\end{abstract}

%%%%%%%%%%%%%%%%%%%%%%%%%%%%%%%%%%%%%%%%%%%%%%%%%%%%%%%%

\section{Introduction}
\label{sec:intro}

\begin{figure}[t]
    \centering
    \subfloat[Multi-hop prompt.]
    {
	   %\label{subfig:multi_hop_demo}
	   \includegraphics[width=0.875\linewidth]{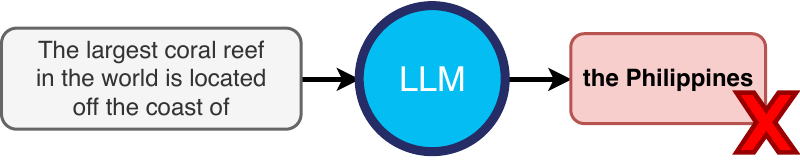} 
    } 

    \subfloat[Multi-hop prompt broken into $2$ single-hop prompts.]
    {
	   %\label{subfig:multi_hop_demo}   
	   \includegraphics[width=0.875\linewidth]{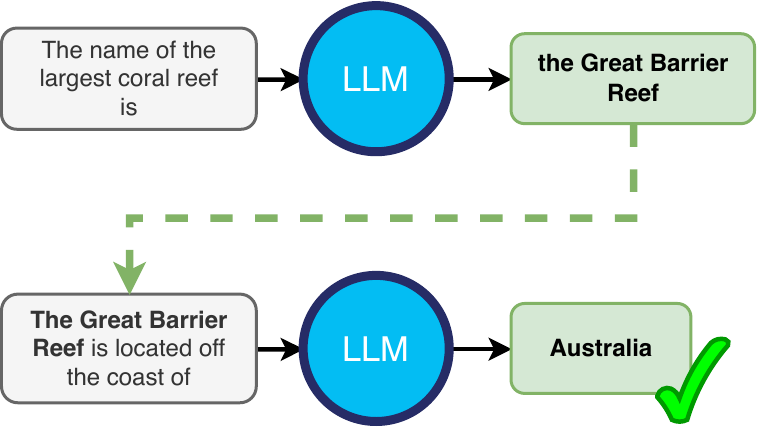} 
    } 
    \caption{A multi-hop prompt vs.\ two analogous single-hop prompts. The outputs are from \gptsmall{}.}
    \label{subfig:multi_hop_demo}
\end{figure}

Transformer-based \emph{Large Language Models} (LLMs) \cite{vaswani2017attention, brown_LanguageModelsAre_2020} 
%are a popular deep learning model for various applications and use cases, from conversational chat-bots (e.g., OpenAI's ChatGPT) to scientific material synthesis pipelines \cite{xie2023large}. While LLMs 
have shown exceptional promise for basic knowledge retrieval and language generation; however, they often lack the ability to perform basic reasoning tasks \cite{arkoudas2023gpt,guo2023indeed,blair2023can}. In this work, we focus on the simple task of answering multi-hop prompts (i.e., prompts in which the subject is not stated explicitly), which humans handle easily but with which LLMs often struggle  (see Fig.~\ref{subfig:multi_hop_demo}). 
%[A prompt is \textit{single-hop} if it states the subject of the relation explicitly, and \textit{multi-hop} otherwise.]
%\mansi{explain figure one more thouroughly and -define what multi- and single- hop prompts are here.}
%TODO

% \begin{figure}[t]
%   \centering
%       \includegraphics[width=0.75\linewidth]{figs/mult-hop-fig_V2.pdf}
%   \caption{Multi-hop reasoning when humans converse.}
%   \label{fig:single-vs-multi}
% \end{figure}

Researchers have attempted to rectify multi-hop reasoning failures by using various prompting methods such as \emph{Chain-of-Thought}~(CoT), \emph{Tree-of-Thought}~(ToT), and \emph{Graph-of-Thought}~(GoT) reasoning \cite{wei_chain-thought_2023,wang2023selfconsistency, tot_long2023large,tot_xie2023decomposition, yao2023tree,besta2023graph}. However, these approaches often put the burden on users to know how to elicit desired responses---and, in the hands of non-expert users, can lead to unreliable prompt completions.
%TODO
% 
Researchers have also proposed model editing \cite{ROME, meng2022mass, mquake, li2023pmet} approaches that may hard-code distant relationships directly into model weights, rather than enhancing the model's abilities to recall and then link simpler relationships. These approaches can be computationally expensive and have unintended effects on other knowledge originally embedded in the model's weights \cite{cohen2023evaluating}. 
%TODO 

Our approach to this problem is based on the hypothesis that LLMs often fail to recall relevant \textit{memories} when attempting to answer a prompt that requires multiple \enquote{hops} of reasoning, rather than lacking knowledge of the \textit{memories} altogether. For example, when attempting to complete the multi-hop prompt, \enquote{The largest coral reef system in the world is located off the coast of\ldots,} we hypothesize that the model does not correctly recall that \enquote{the largest coral reef system in the world} is \enquote{the Great Barrier Reef} before predicting the next token in the sequence. 
%Therefore, we posit that the model does not incorporate the relevant information to produce an accurate next token prediction.
Yet the model can accurately complete both the corresponding single-hop prompt \enquote{The Great Barrier Reef is located of the coast of\ldots,} and, when prompted,  \enquote{the largest coral reef} as \enquote{the Great Barrier Reef.} Clearly, this information was encoded in the model during training but is not incorporated when answering questions that reference the prompt's subject indirectly. In this case, therefore, we  define the missing \emph{memory} to be \enquote{the Great Barrier Reef.} 
%Model's have to recall fewer \textit{memories} for single-hop prompts before predicting the next token. 

%rather than lacking the information altogether. Instead of editing the weights of a model to re-encode redundant relationships or employing a complicated prompting paradigm, we simply intervene on an inference pass and inject useful information, or "memories", into key layers to refocus the model's train-of-thought as it attempts to complete a multi-hop reasoning prompt.
To study our hypothesis, we first attempt to reverse engineer a key mechanism by which transformer-based LLMs conduct reasoning. Specifically, we find that in transformer-based models it is attention heads, rather than multi-layer perceptrons, that are responsible for retrieving \emph{memories} critical to successful model predictions; our finding is further substantiated by similar findings by \citet{li2023pmet,geva_dissecting_2023,dar2022analyzing}. We then study instances in which this mechanism fails in multi-hop reasoning tasks and find that this mechanism is likely the source of incorrect, insufficient, or irrelevant \textit{memory} retrievals \textbf{(Contribution 1)}---for an example, see Fig.~\ref{fig:transformer diagram}.

%\mansi{Everything following this can be compressed a little. I don't think we need to be as technical in the intro if we set the stage properly for what our main problem is and if we defined all relevant new vocabulary.}
%% Probably rephrase. This is very jargon heavy.
We then propose a lightweight \emph{memory injection} method that can be employed to correct a multi-hop reasoning failure during inference \textbf{(Contribution~2)}. As an example: by employing our method to inject the \emph{memory} of \enquote{The Great Barrier Reef} into the multi-hop prompt \enquote{The largest coral reef system in the world is located off the coast of\ldots} during inference, we increase the probability of the next token \enquote{Australia} by $189\%$; refer to Fig.~\ref{fig:injection_mechanism} for details. 

For our analyses, we hand-crafted a dataset for interpretabilty purposes \textbf{(Contribution 3)} and make use of a larger programmatically-generated dataset---refer Table~\ref{tab:data} for more information.

Finally  we conduct additional experiments \textbf{(Contribution 4)} to:
\begin{enumerate}[itemsep=-.25ex]
    \item Identify the ideal layer and magnitude for the \textit{memory injection}.
    %\item Ensure that our \emph{memory injections} are robust to random noise.
    \item Demonstrate the significance of curating prompt-specific \textit{memories} for injection.
    \item Analyze if \textit{memories} drawn from different parts of speech---namely, nouns, adjectives, adverbs, conjunctions, verbs---behave differently during \textit{memory injection}.
\end{enumerate}

\section{Background \& Notation}
\label{sec:background}
We define %clarify the distinction between 
single- vs.\ multi-hop prompts and provide a formal definition of the transformer model.

\subsection{Multi-hop vs. single-hop prompts}
%We seek to compare and contrast how LLMs perform next-token prediction when given an explicit (single-hop) vs.\ ambiguous (multi-hop) prompt. 
% The ultimate goal is to identify mechanisms to improve prediction on more ambiguous prompts. 
% For the sake of completeness, here we clarify what we mean by single-hop and multi-hop.
% We want to compare and contrast the prediction mechanisms employed by LLMs when given a single-hop prompt and its multi-hop counterpart. For the sake of completeness, we define and clarify the difference between single-hop and multi-hop prompts.

We refer to a prompt as \textit{single-hop} if the subject of the relation is stated explicitly in the prompt, and \textit{multi-hop} otherwise. Multi-hop prompts refer to their subject in a way that requires an additional \enquote{hop} or inference step. For example, consider the single-hop prompt, \enquote{\emph{George Washington} fought in the\ldots} with a correct answer being \enquote{Revolutionary War.} In the analogous multi-hop prompt, \enquote{\emph{The first president of the United States} fought in the\ldots,} a preliminary inference step is needed to identity of the first US president before predicting the next token. For additional examples of single- and mutli-hop prompts, see Table~\ref{tab:prompt_examples} in the appendix.

\subsection{Transformer Architecture}
We introduce a common notation for the components of the transformer-based language model architectures that are the focus of our analyses. 
Specifically, we focus on auto-regressive, decoder-only models. 
We adopt much of our notation from \citet{elhage2021mathematical} and \citet{geva_dissecting_2023}.

\subsubsection{Embedding Inputs}
An input text is parsed into $N$ distinct tokens $t_0, \cdots, t_N$. Each token $t_i$ is then embedded as $x_{i}^{0} \in \mathbb{R}^d $ via an embedding matrix $ W_{E}  \in  \mathbb{R}^{|V| \times d}$, where $V$ is the vocabulary and $d$ is the hidden dimension.

\subsubsection{Residual Stream}
Following the embedding layer, all tokenized embeddings $x_i^0$ are passed through a series of \emph{residual blocks}. The outputs of each \emph{residual block} are added back into the model's \emph{residual stream} denoted by $R^{\ell}\;(\forall \ell\in\{1,\cdots,L\}
)$ where $L$ is the number of layers in the LLM.

We define the \emph{residual stream} at layer $\ell$ as:
\begin{equation}
    R^{\ell} = [x^{\ell}_0, \cdots, x^{\ell}_N], 
\end{equation}

\noindent
where $x^{\ell}_i$ is the representation of token $i$ at layer~$\ell$. The \emph{residual stream} is updated by its respective \textit{residual block} $r^{\ell}$:

\begin{equation}
    R^{\ell+1} = R^{\ell} + r^{\ell+1},
    %R^{\ell+1} = R^{\ell} + a^{\ell+1}+ m^{\ell+1},
\end{equation}

\noindent
and the output of a \textit{residual block} $r^{\ell}$ is:
\begin{equation}
     r^{\ell} = a^{\ell}+ m^{\ell},
\end{equation}

\noindent
where $a^{\ell}$ is the output of the \emph{Multi-Headed Self Attention}~(MHSA) layer and $m^{\ell}$ is the output of the \emph{Multi-Layer Perceptron}~(MLP). We define MHSA and MLP in the following sections.

\subsubsection{Multi-Headed Self Attention (MHSA)}
Each MHSA layer $\ell$ is defined via four parameter matrices $W^{\ell}_Q, W^{\ell}_K, W^{\ell}_V, W^{\ell}_O \in \mathbb{R}^{d \times d}\;(\forall \ell\in\{1,\cdots,L\})$ and the hyperparameter $H$ denotes the number of attention heads.
Following \citet{elhage2021mathematical} and \citet{geva_dissecting_2023}, we can further dissect our parameter matrices to better observe the relationship between unique sets of parameters and individual attention heads: 
$W_Q^{l,j}, W_K^{\ell,j}, W_V^{\ell,j} \in \mathbb{R}^{d \times \frac{d}{H}}$ and $W_O^{\ell,j} \in \mathbb{R}^{\frac{d}{H} \times d}$ for $j \in [1,H]$. Now, we can define the output of each MHSA $a^{\ell}$ as the sum of all attention head outputs,

\begin{equation}
    a^{\ell} = \sum_{j=1}^H h^{\ell,j},
\label{eq:mhsa_output}
\end{equation}

\noindent
where $h^{\ell,j}$ is the output of the $j^{th}$ head in layer $\ell$:
%Its formal definition is defined below:

\begin{equation}
    h^{\ell,j} = A^{\ell, j}
    \big(R^{\ell-1} W_V^{\ell, j}\big) W_O^{\ell, j}.
\label{eq:attention_output}
\end{equation}

% \nathaniel{I would add a sentence here along the lines of, \enquote{where $A^{\ell,j}$ is the...,} to make the notation more accessible.}
% $A^{\ell,j}$ is defined below,

\begin{equation}
\small
    A^{\ell, j}= \softmax 
    \Bigg(
        \frac
        {\big(R^{\ell-1}W_Q^{\ell,j}\big) \big(R^{\ell-1}W_K^{\ell,j}\big)^T}
        {\sqrt{\sfrac{d}{H}}} 
         \odot M
    \Bigg)
\label{eq:attention}
\end{equation}

\noindent 
where the $\softmax(\cdot)$ is performed as a row-wise operation, $\odot$ is the Hadamard product, and $M \in \{0, 1\}^{N \times N}$ is an auto-regressive attention mask where masked token positions are set to $0$. %, and $h^{\ell,j}$ is the output of the $j^{th}$ attention head in layer $\ell$. 

\subsubsection{Multi-Layer Perceptron (MLP)}
Each MLP is defined via two parameter matrices $W^{\ell}_F, W^{\ell}_I \in \mathbb{R}^{d \times d_p}$ with inner-dimension $d_p$ and a nonlinear activation function, $\sigma$. 

\begin{equation}
    m^{\ell} = W^{\ell}_F\; 
    \sigma
    \Big(
        W^{\ell}_I \big(a^{\ell} + R^{\ell-1}\big)
    \Big)
\end{equation}

%\mansi{Elaborate on MLPs as Key-values stores here.}

\subsubsection{Unembedding Predictions into Logits}
After the final \textit{residual block}, all token positions $x_i^{-1}$ will be projected back into the vocabulary domain via the \emph{unembedding matrix} $W_{U} \in  \mathbb{R}^{d \times |V|}$. The output of the last token position is the next token prediction of the model.

%%%%%%%%%%%%%%%%%%%%%%%%%%%%%%%%%%%%%%%%%%%%%%%%%%%%%%%%%%%%%%%%%%%%%%%%%%%%%%%%
%%%%%%%%%%%%%%%%%%%%%%%%%%%%%%%%%%%%%%%%%%%%%%%%%%%%%%%%%%%%%%%%%%%%%%%%%%%%%%%%

\section{Experimental Overview}
\label{sec:experiments}
% For this work, 
Our central aim is to better understand how the outputs of the attention heads affect model performance with respect to predicting the correct next token in prompts requiring single-hop reasoning versus in prompts requiring multi-hop reasoning. 
%We hypothesize that multi-hop prompts prove more difficult to accurately complete since they require a model to conduct additional reasoning to arrive at the correct answer. To run our experiments, we use various tools to analyze attention heads and their outputs.
% we now aim to formally investigate if poorer model performance on multi-hop prompts is due to its inability to resolve the intermediate \enquote{hop}.

\subsection{Dataset Descriptions}

We employ three datasets in this work. Two, used to assess model prompt completion accuracy, are our own high-quality manually curated dataset of single and multi-hop pairs and a programmatically generated dataset of prompt pairs.
%see Tables~\ref{tab:data} and~\ref{tab:prompt_examples} and below. %Additionally, we measure the subsequent improvement or degradation in performance on these datasets after applying a novel intervention detailed in Section~\ref{subsec:correcting_multihop_failures} designed to improve model performance by injecting relevant information into specific attention heads. 
The third comprises lists of words from common parts of speech, which 
%we also use a large word frequency dataset to generate lists of words in
%the most 
%common parts of speech.
we use to study how the effectiveness of our intervention varies with the part of speech of injected tokens.
%we also use a large word frequency dataset to generate lists of words in
%the most 
%common parts of speech. % in American English.

\subsubsection{Programmatically Generated Dataset}
The 2WikiMultiHop~dataset~\cite{2wmh} contains pairs of knowledge triples $\{(s_1, r_1, s_2)_1, (s_2, r_2, s_3)_2\}$, each with two subjects $s$ and a relationship $r$.
We used these knowledge triples, plus a set of predefined templates, to generate a set of pairs of single- and multiple-hop questions, \harddata{}\/: see Tables~\ref{tab:data} and~\ref{tab:prompt_examples}. 

For example, let
$ s_1$ =  \enquote{Lilli's Marriage,} $r_1 = $\enquote{director,} $s_2$ = \enquote{Jaap Speyer,} $r_2 =$ \enquote{country of citizenship,} $s_3$ = \enquote{Dutch.} Then for
\textbf{single-hop}, the template: \enquote{The $r_2$ of $s_2$ is \ldots $s_3$}, the prompt yields the prompt 
\enquote{The country of citizenship of Jaap Speyer is \ldots [Dutch]}; for 
\textbf{multi-hop}, the template \enquote{The {$r_2$} of the $r_1$ of $s_1$ is \ldots $s_3$} yields then the prompt:
\enquote{The country of citizenship of the director of Lilli's Marriage is \ldots [Dutch].}

%\mansi{Space permitting, maybe touch on the fact that 2wmh is harder than Hand?}

\subsubsection{Human-Generated Dataset}
As evidenced by the example presented above, the \harddata{} dataset, while scalable, contains many grammatical flaws. Therefore, we construct an additional dataset for multi-hop reasoning with a focus on grammatical and factual correctness presented below.
We hand-crafted 106 (single-hop, multiple-hop) prompt pairs, each in the same form as those in \harddata{}: e.g.,
%We constructed this dataset by using the same framework as \harddata{}, example from this dataset is:
%\noindent
\textbf{single-hop}:
\enquote{St. Peter's Basilica is in the city of\ldots [Rome]} and 
\textbf{multi-hop:}
\enquote{The biggest church in the world is in the city of\ldots [Rome]}.
Each prompt pair was also evaluated by two external reviewers for factual and grammatical accuracy.
We hereafter refer to this dataset as \easydata{}; see Tables~\ref{tab:data} and~\ref{tab:prompt_examples}.

%The single-hop and multi-hop prompts are semantically identical except that the single-hop prompt possesses an explicit subject (e.g., \enquote{George Washington}) while the multi-hop prompt contains an obscure subject that corresponds to the explicit subject (e.g., \enquote{The first president of the United States}). 
% We refer to this dataset as \enquote{Hand.}
%\mansi{Give example of prompts}

\begin{table*}[hbt!]
\small
\centering
\begin{tabular}{cccllllll}
\toprule
    \multicolumn{3}{}{} & \multicolumn{3}{c}{Single-hop} & \multicolumn{3}{c}{Multi-hop}
\\
\cmidrule(lr){4-6}
\cmidrule(lr){7-9}
    Data & Size & Model & Answer prob. & Surprisal & Prompt len. & Answer prob. & Surprisal & Prompt len.
\\
\midrule 
   \easydata{} & 106 & \gptsmall{} & 0.157 & 4.21 & 9.66 & 0.087 & 4.91 & 12.99 
\\

    \easydata{} & 106 & \gptlarge{} & 0.28 & 2.90 & 9.66 & 0.157 & 3.97 & 12.99 
\\

    \harddata{} & 1000 & \gptsmall{} & 0.0007 & 9.80 & 10.44 & 0.00086 & 9.64 & 14.00 
\\

    \harddata{} & 1000 & \gptlarge{} & 0.0023 & 8.71 & 10.44 & 0.002 & 8.57 & 14.00 
\\
\bottomrule
\end{tabular}
\caption{Properties of the datasets used in our work. \emph{Size}: Number of prompts. \emph{Answer prob.}: Average model probability model for expected next token. \emph{Surprisal}: Average model surprisal value for expected next token ($\mathit{surprisal} \triangleq -\log(p)$ where $p$ is a probability). \emph{Prompt len.}: Average tokenized length of prompt.}
\label{tab:data}
\end{table*}

\subsubsection{Part of Speech Dataset}

We used a subset of the Corpus of Contemporary American English \cite{davies2011word} which compiles word frequencies \cite{davies2010corpus} to generate lists of (i) the most common words from various parts of speech: 824 adjectives, 331 adverbs, 40 conjunctions, 2635 nouns, 969 verbs, and (ii) the 5050 most common words overall (``top 5050'').

%We used the dataset by \citet{davies2010corpus} containing word frequencies from the Corpus of Contemporary American English \cite{davies2011word} to generate lists of (i) the most common words from various parts of speech: 824 adjectives, 331 adverbs, 40 conjunctions, 2635 nouns, 969 verbs, and (ii) the 5050 most common words overall (``top 5050'').

\subsection{Model Description}
We work with two pretrained GPT2 models \cite{radford_llmsOpenAI_2019}. 
\textbf{\gptsmall{}} has 12 layers, 12 attention heads per attention layer, and $\sim$160M parameters.
\textbf{\gptlarge{}} has 36 layers, 20 attention heads per attention layer, and $\sim$840M parameters.
Both have a vocabulary of $\sim$50K tokens.

\subsection{Tools \& System Setup}
We use the \textit{Transformer Lens} Python package \cite{transformer_lens} to cache, inspect, and construct interventions on model inference passes. 
We ran experiments on a single A100 GPU with $40$~GB RAM.
Experimental code, dependency information, and datasets are available on GitHub.\footnote{\url{https://github.com/msakarvadia/memory_injections}}
%\footnote{Anonymized for double-blind review}

\section{Proposed Methods}
%\section{Proposed Model Inspection and Memory Injection Method}
\label{sec:algorithm}
Recent work %by \citet{geva_dissecting_2023} 
suggests that attention heads are knowledge retrievers during a model's inference pass \cite{geva_dissecting_2023, li2023pmet}. Extending this result to multi-hop prompts, we hypothesize that attention layers play an important role in retrieving memories relevant to the ``hop'' in a given prompt. Therefore we define two algorithms below: one for analyzing attention head outputs in embedding space and the other for injecting a targeted memory into a model's hidden activations in order to correct faulty/incomplete reasoning.

%for determining the best location in the model's architecture to embed a targeted memory in order to correct faulty/incomplete reasoning.

\subsection{Interpreting Attention Heads}

We want to further understand the outputs of individual heads, and more specifically assess if any individual attention heads are exercised differently by single-hop vs. multi-hop prompts. 

Inspired by Logit Lens \cite{logitlens}, we leverage the model's unembedding matrix to  %gain insight into 
study the internal mechanism of each attention head. For attention head $j$ in layer $\ell$, $h^{\ell,j}$, we apply the model's unembedding matrix $W_U$ followed by a $\softmax(\cdot)$ operation % (denoted by $\lambda$) 
and interpret the last token position (out of $N$ total tokens) as a set of probabilities over tokens in the vocabulary space: %, for $N$ tokens:
\begin{equation}
     \mathit{vocab}^{\ell,j} = \softmax(h^{\ell,j} W_U )_{N-1} 
\label{eq:interp_attn}
\end{equation}

\begin{figure}[t]
  \centering
      \includegraphics[width=\linewidth]{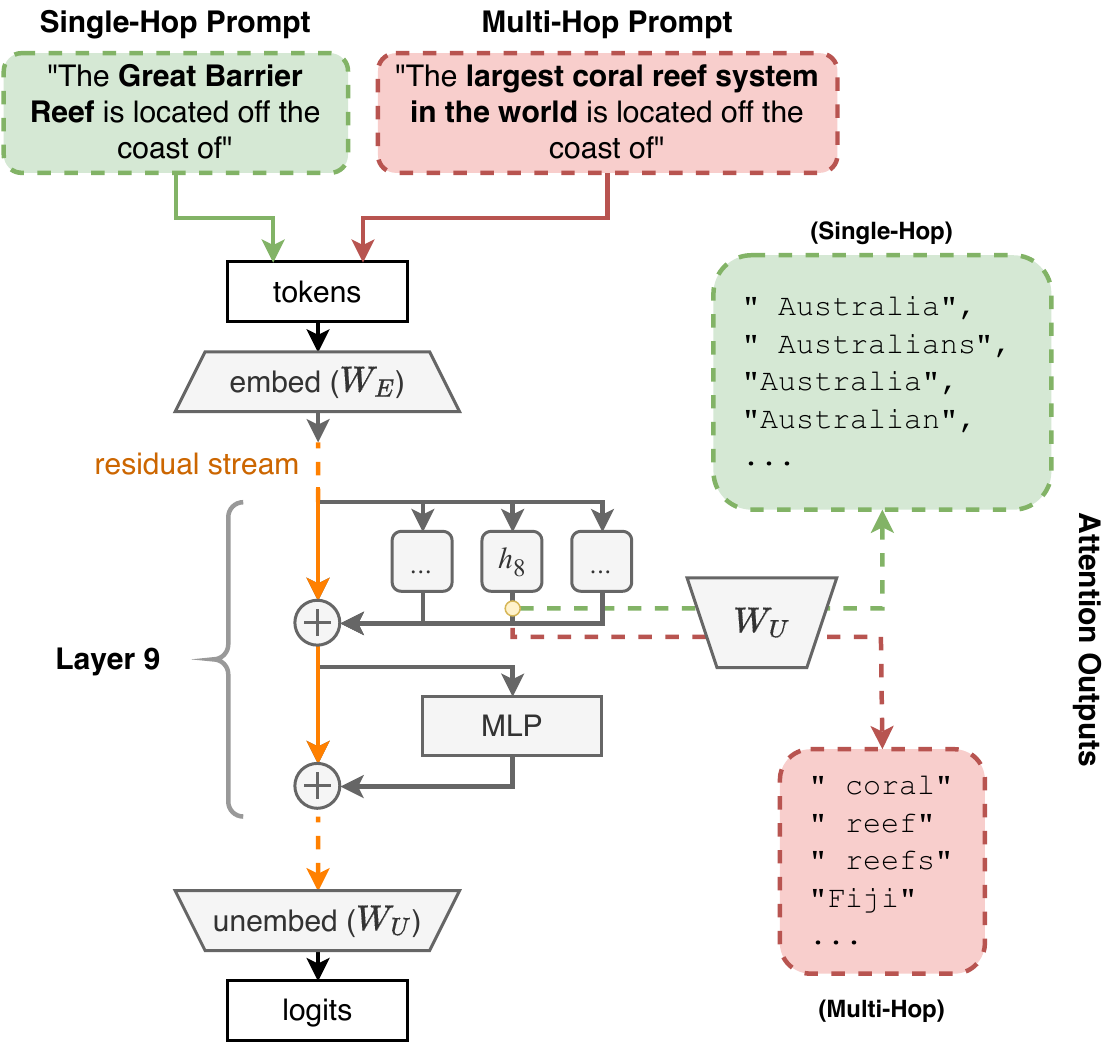}
  \caption{
        \textbf{Diagram of language model reasoning.} 
        Highest ranked attention outputs of \gptsmall{} at layer $\ell=9$, head $h=8$ when projected into vocabulary space (via the \gptsmall{} embedding matrix) for a single-hop prompt (green) and its multi-hop counterpart (red).
        %Attention outputs of \gptsmall{}  at layer $\ell=9$, head $h=8$ being projected into vocabulary space (via the \gptsmall{} embedding matrix) for a single-hop prompt and its multi-hop counterpart.
    }
  \label{fig:transformer diagram}
\end{figure}

\noindent
%where $N$ is the number of tokens. 
See in Fig.~\ref{fig:transformer diagram} an example of discrepancy in attention head behavior, when using Eq.~\eqref{eq:interp_attn}, for analogous single vs.\ multi-hop prompts. See additional examples in Table~\ref{tab:atten_head_outputs}.

A potential limitation of this approach is that it may portray attention head behavior inaccurately due to %the presence of 
representational drift between model layers---and, like  \cite{logitlens}, may not generalize to other models. Nevertheless, we find it to be an effective preliminary tool for studying the function of attention heads in updating the output distribution. We leave the development of an interpretability tool that considers these drawbacks to future work.

%\mansi{[Elaborate on the fact that we understand that applying the unembedding matrix to intermediate layers may not be the best approach but for now it seems to work. Additionally, we aim to address this shortcoming in future work.]}

%\mansi{[Add a table here with examples of outputs from specific attention heads that greatly differs between various single and multi hop prompts]}

\subsection{Memory Injections to Correct %Multi-hop Reasoning 
Failures} \label{subsec:correcting_multihop_failures}
Fig.~\ref{fig:transformer diagram} shows how Eq.~\eqref{eq:interp_attn} can reveal discrepancies between attention head behaviors for single- vs.\ multi-hop prompts.
%We use Eq.~\eqref{eq:interp_attn} to identify discrepancies between attention head behaviors, such as that shown in Fig.~\ref{fig:transformer diagram}. 
We hypothesize that such discrepancies 
arise because the model, when updating the output distribution in each layer, fails to 
incorporate information about the implicit entity in the multi-hop prompt. This seems reasonable, as to retrieve information about an implicit entity one likely must first relate that entity to some explicit subject and then retrieve relevant information (hence our notion that processing prompts with implicit subjects requires an extra hop compared to those with explicit subjects).
% to reason about what the \enquote{hop} for a multi-hop prompt. 

\begin{figure}[t]
  \centering
    \includegraphics[width=\linewidth]{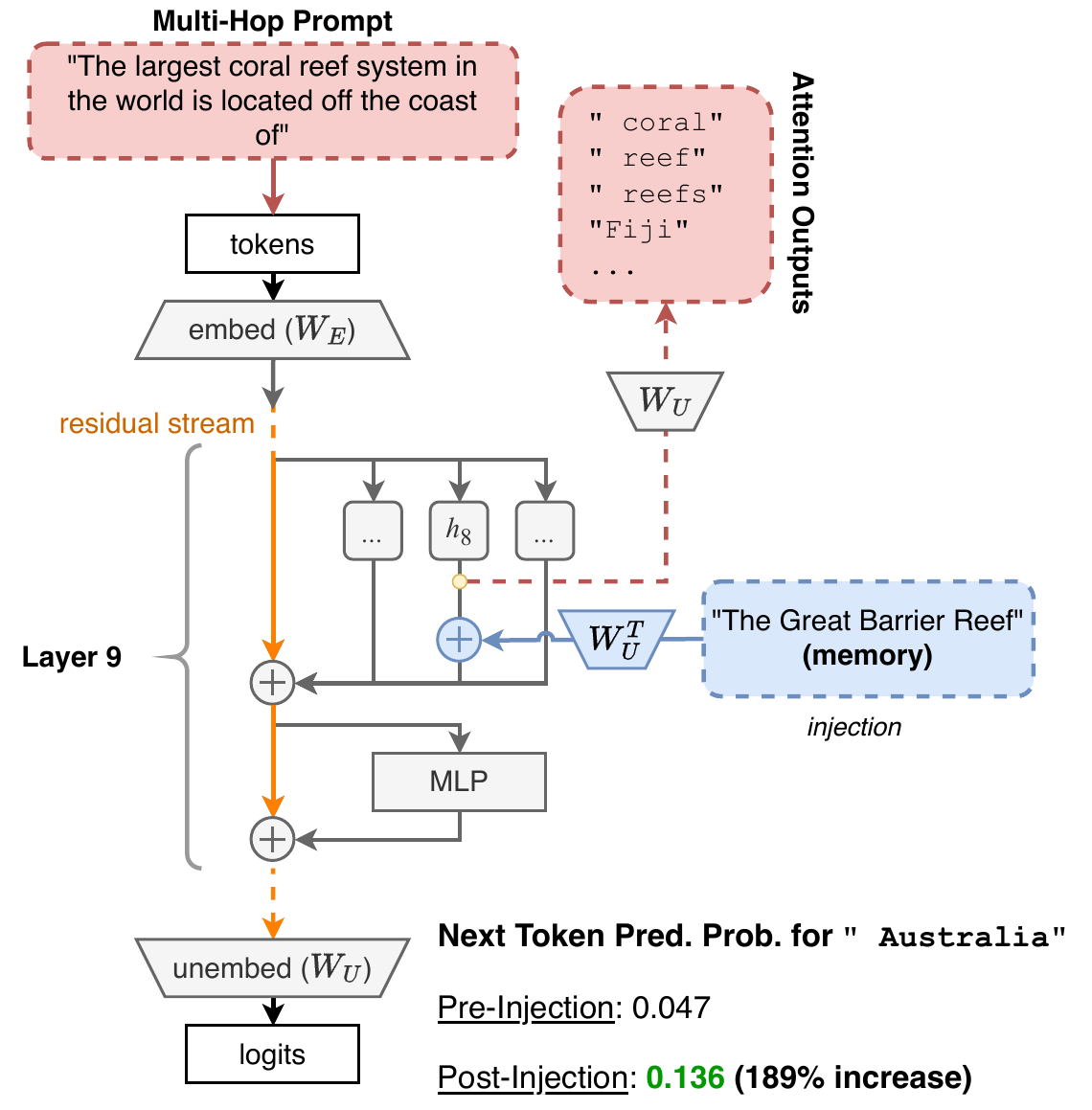}
    \caption{
        \textbf{Memory injection.} Injecting memory \enquote{The Great Barrier Reef} into \gptsmall{} hidden activations at layer $\ell=9$, head $8$, $\tau=4$.
    }
  \label{fig:injection_mechanism}
\end{figure}

Thus we design a 
%simple and efficient 
method (see Fig.~\ref{fig:injection_mechanism}) for injecting a missing hop directly into the output hidden states of an attention head
%$h^{l,j}$ where $l$ is the layer number and $j$ is the head number 
before those outputs are added back into the transformer's residual stream:
%We detail the algorithm below:
\begin{enumerate}
    \item Let $m$ be a memory (a phrase, for example: \enquote{The Great Barrier Reef}) and let $\tau$ be the magnitude of the memory injection.
    %(we also refer to this quantity, $\tau$, as the \textit{tweak factor}). 

    %\item Tokenize the memory $m$ into $t_0,\cdots,t_q$ where $q$ is the number of tokens and each $t_i$
    \item Tokenize the memory $m$ into $t_0,\cdots,t_q$ where $q$ is the number of tokens. We encode each token $t_i$ into a one-hot vector $b_i \in \{0,1\}^{|V|}$ and sum all resulting one-hot vectors $b_i$ together into a binary vector $B\triangleq \sum_{i} b_{i}$. 
    
    \item Embed the binary vector, $B$, back into the model's latent space by applying the transpose of the unembedding matrix:
    \begin{equation}
        B^* = B\, W_U^T
    \end{equation}
    
    \item Then, to inject a memory at the attention layer of layer~$\ell$, add the embedded memory into the outputs of the attention heads during the inference pass:
    \begin{equation}
        a^{\ell} = \sum_{j=1}^H h^{\ell,j} + \tau  B^*
    \end{equation}
    
\end{enumerate}

\noindent
See additional examples of \textit{memory injections} in Table~\ref{tab:memory_injections}.

\section{Results and Discussion}
\label{sec:results}

We report, in turn, on our curated memory, random memory, and part-of-speech injection experiments.

\subsection{Curated Memory Injections}
\label{subsec:curated_mem_injec}

\begin{figure*}[t]
  \centering
  \includegraphics[width=\linewidth]{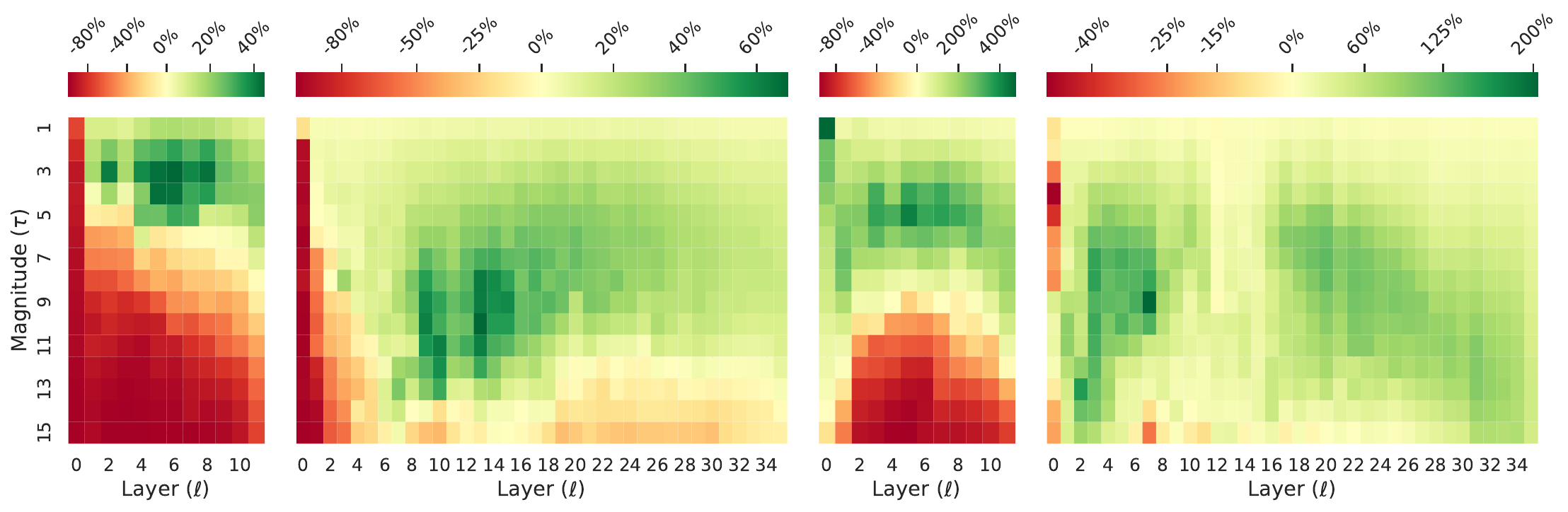}
  \caption{\textbf{Curated memory injections.} From left to right: \gptsmall{} + \easydata{}, \gptlarge{} + \easydata{}, \gptsmall{} + \harddata{}, \gptlarge{} + \harddata{}. Each cell in each heatmap is the average percent difference between the pre- and post-injection next token predictions for multi-hop prompts. Green cells denote a positive percent difference (i.e., correct prediction is more likely), while red cells denote a negative percent difference (i.e., correct prediction is less likely). When computing the averages for each 
  ($\ell$, $\tau$) pair we exclude outliers not within $\pm2$ standard deviations from the mean.
  }
  \label{fig:curated_memories}
\end{figure*}

We hypothesize that a model's poor performance on multi-hop prompts is due to its inability to resolve the implicit subject (e.g., \enquote{The largest coral reef system in the world}) to an explicit subject (e.g., \enquote{The Great Barrier Reef}). This failure limits the later layers' ability to retrieve relevant information about this subject before predicting the next token. Therefore, in this experiment, we curate sets of tokens to inject into our model's residual stream such that it can resolve the explicit subject more easily. We further study the effect that the injection magnitude $\tau$ has on its success.

\textbf{Experimental design:}
For every multi-hop prompt in our datasets, we extract the explicitly stated subject from the corresponding single-hop prompt and inject those tokens as \emph{memories} into each attention layer as described in Section~\ref{subsec:correcting_multihop_failures}. For example, given the
\textbf{single-hop prompt} \enquote{\textit{The Great Barrier Reef} is located off the coast of\ldots} and the
\textbf{multi-hop prompt} \enquote{\textit{The largest coral reef system in the world} is located off the coast of\ldots,} the \textbf{memory} is
\enquote{\textit{The Great Barrier Reef.}}

%\begin{tabbing}
%aa\=aa\=\kill
%\>Single-hop prompt: ``\textit{George Washington} fought\\
%\>\>\textit{in the \ldots}''\\
%\>Multi-hop prompt: ``\textit{The first president of the}\\https://www.overleaf.com/project/6477ad31b5ce86f40bc029d2
%\>\>\textit{United States} fought in the \ldots''\\
%\>Injection: \enquote{\textit{George Washington}}
%\end{tabbing}

%\\Single-hop prompt: \enquote{\textit{George Washington} fought in the ...}
%\\Multi-hop prompt: \enquote{\textit{The first president of the United States} fought in the ...}
%\\Memory: \enquote{\textit{George Washington}}

We assess the effects of injection layer $\ell$ and magnitude $\tau \in [1, \cdots, 15]$ by enumerating the resulting change in accuracy for all combinations of these two parameters for both \gptsmall{} and \gptlarge{}.
We measure the success of a memory injection by calculating the percent increase between the model's predicted probability for the expected next token from the multi-hop prompt with and without the injection. A greater positive difference indicates a more successful injection.

\textbf{Discussion:}
Results are in Fig.~\ref{fig:curated_memories}. We observe that each model/dataset combination has an optimal layer $\ell$ and magnitude $\tau$ for memory injections: the darkest green areas, which signify the highest average percent increase in probability of the expected next token for the respective dataset. The best ($\ell$, $\tau$) pair injection results are in Table~\ref{tab:curated_vs_random}. Additional examples of memory injections are in Table~\ref{tab:memory_injections}.

%\mansi{Need to comment further on what these graphs tell us about the trends between layer number/magnitude overall for our datasets.}

\begin{table*}[hbt!]
\small
\centering
\begin{tabular}{cccc|rrrrrrr}
\toprule
%     \multicolumn{4}{}{} & \multicolumn{7}{c}{Memory Injection}
% \\
% \cmidrule(lr){5-11}
    \multicolumn{4}{}{} & Curated & \multicolumn{6}{c}{Random}
\\

\cmidrule(lr){5-5}
\cmidrule(lr){6-11}
    Model & Data & $\ell$ & $\tau$ &  Subject & Adj. & Adv. & Conj. & Noun & Verb & Top-$5050$ 
\\
\midrule 
    GPT2 Small & Hand & 7 & 3 & \textbf{45\%} & -7.6\% & -6.0\% & -6.3\% & -6.5\% & -7.5\% & -6.0\%
    %GPT2 Small & Hand & 7 & 3 & 53\% & -8.3\% & -8.7\% & -6.2\% & -7.2\% & -7.5\% & -5.9\%
\\
%\midrule 
    GPT2 Small & 2wmh & 6 & 5 & \textbf{424\%} & -17.1\% & -15.1\% & -10.3\% & -1.1\% & -1.2\% & 1.6\%
    %GPT2 Small & 2wmh & 8 & 4 & 319\% & -10.5\% & -11.4\% & -9.0\% & -8.9\% & -9.4\% & -7.5\%
\\
%\midrule 
    GPT2 Large & Hand & 14 & 10 & \textbf{68\%} & -8.1\% & -4.4\% & -4.9\% & -9.8\% & -6.0\% & -4.7\%
    %GPT2 Large & Hand & 14 & 9 & 70\% & -7.1\% & -7.3\% & -2.3\% & -6.6\% & -4.2\% & -3.8\%
\\
%\midrule 
    GPT2 Large & 2wmh & 8 & 9 & \textbf{204\%} & 13.0\% & 11.6\% & 3.5\% & 11.8\% & 4.3\% & 17.6\%
    %GPT2 Large & 2wmh & 4 & 8 & 161\% & -6.4\% & -7.5\% & -3.1\% & -5.1\% & -9.4\% & -3.8\%
\\
\bottomrule
\end{tabular}
\caption{\textbf{Curated vs.\ random memory injections.} Table shows the ($\ell$, $\tau$) pairs for the best token injections, along with the \emph{average percent difference} (excluding outliers >$\pm2$ standard deviations from the mean) between pre- and post-injection expected next token predictions for multi-hop prompts. Each random injection column indicates 40 random injections from [Adjectives, Adverbs, Conjunctions, Nouns, Verbs, Top 5050] at the ideal ($\ell$, $\tau$).}
\label{tab:curated_vs_random}
\end{table*}

\subsection{Random Memory Injections}
In Section~\ref{subsec:curated_mem_injec}, we identify ideal ($\ell$, $\tau$) pairs for each model and dataset for a curated memory injection. We now demonstrate that the results we observe are not spurious: i.e., the information that we inject at each head should be related to the explicit subject. We demonstrate the need for our particular injection routine by assessing the effects on model accuracy of randomly injecting tokens from various parts of speech.

\textbf{Experimental design:} 
We conduct targeted injections for the high-scoring ($\ell$, $\tau$) pairs identified via the experiment in Section~\ref{subsec:curated_mem_injec}, Table~\ref{tab:curated_vs_random}. 
Instead of injecting curated subject tokens, we select as candidate injections the 40 most common words from each of the adjectives, adverbs, conjunctions, nouns, verbs, and top 5050 subsets of our \textit{Part of Speech} dataset.
We then apply each word as an individual injection for every prompt in our multi-hop dataset at the ideal ($\ell$, $\tau$) pair. We term these injections \enquote{random,} as they were not curated to be relevant to our prompts.

\textbf{Discussion:}
The results are in the right half of Table~\ref{tab:curated_vs_random}. 
We observe that a random injection led, on average, to a degradation in predictive performance across most parts of speech considered, as indicated by a negative percent difference (decrease in correct answer probability) between the pre- and post-injection expected next token probabilities for multi-hop prompt completions. Additionally, no random injection result exceeded the performance of a curated injection.
These findings suggest that the choice of injected tokens is critical for improving multi-hop prompt completion success.

\subsection{Memory Injections for Parts of Speech}
\begin{figure*}[t]
  \centering
      \includegraphics[width=\textwidth]{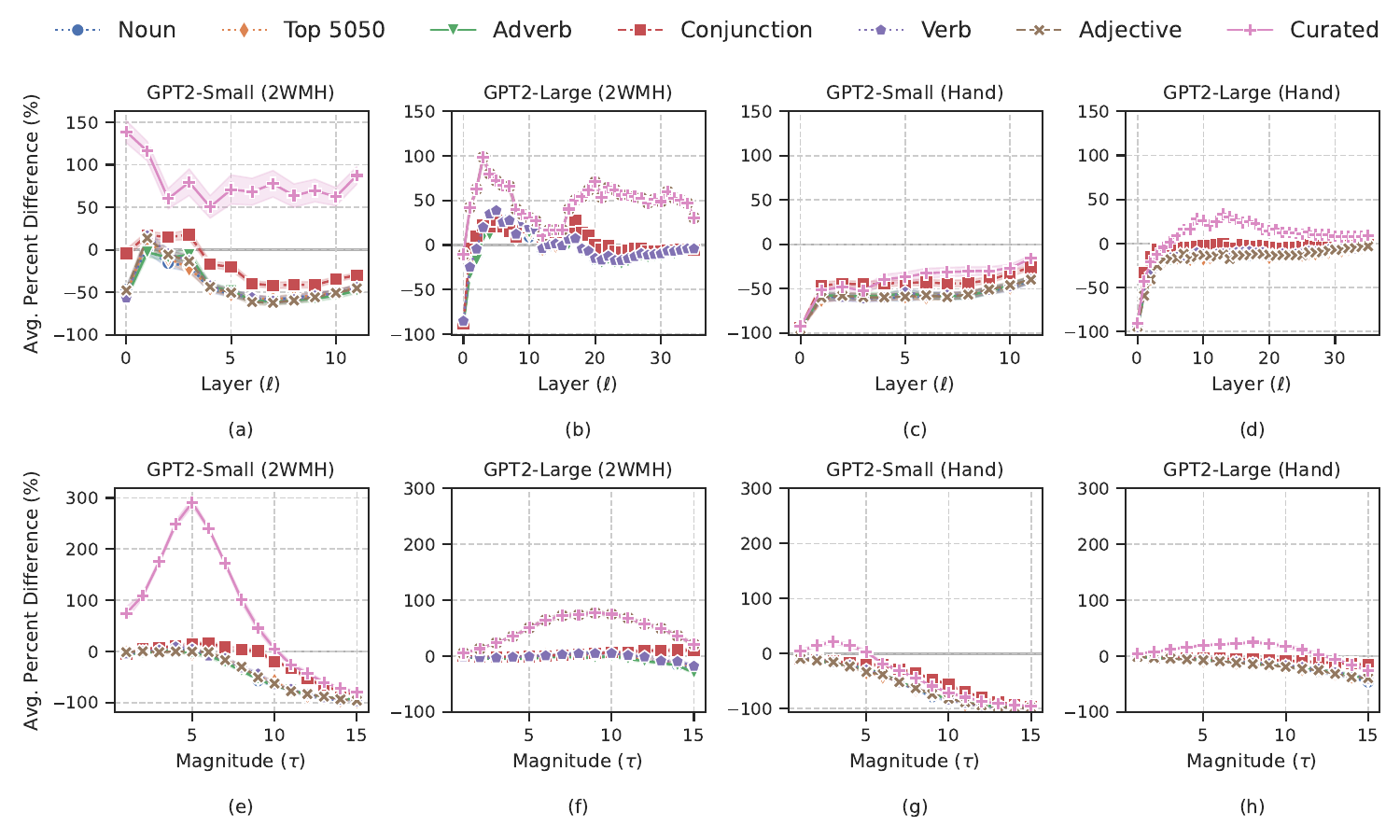}
  \caption{\textbf{Part of speech memory injections.} This figure shows the average effect of memory injections from various parts of speech as a function of layer $\ell$ (top row) and magnitude $\tau$ (bottom row). The standard deviation scaled by 10\% is pictured across magnitudes (top row) and layers (bottom row).}
  \label{fig:pos_result}
\end{figure*}

We have tested curated vs. random memory injections at ideal ($\ell$, $\tau$) pairs. Now we assess whether memory injections from specific parts of speech more broadly have positive impacts on prompt completions, not just at the ideal locations for curated memories, but also at other ($\ell$, $\tau$) pairs.
Our hypothesis is that if a transformer-based LLM has learned a division of labor regarding which attention layers are responsible for retrieving specific concepts (e.g., parts of speech) then this experiment might highlight those learned roles.

\textbf{Experimental design:} 
This experiment is identical to that of
%Our methods are as in
Section~\ref{subsec:curated_mem_injec},
except that: (i) for
%however, the method by which we construct our injected tokens differs. For 
each part of speech $\mathit{pos} \in$ [adjectives, adverbs, conjunctions, nouns, verbs, top 5050], we use a randomly selected word:
e.g., \enquote{apple} from \enquote{nouns}; and (ii)
%from it and use it 
%as the injection. For example, if the part of speech was \enquote{nouns,} the randomly selected injection might be \enquote{apple.}
when searching for the ideal ($\ell$, $\tau$) pair for a given part of speech and  multi-hop prompt, we use a new random word for each injection.
%every injection uses a new random word. Meanwhile, in Section~\ref{subsec:curated_mem_injec}, the constructed token injection was held constant for each multi-hop prompt. We conduct this search for all parts of speech.

\textbf{Discussion:}
The results are in Fig.~\ref{fig:pos_result}. We note that for no part of speech considered here does the average performance of the studied memory injections exceed that of the curated memory injections presented in Table~\ref{tab:curated_vs_random}. Additionally, memory injections from adjectives, adverbs, nouns, verbs, and top 5050 seemed to exhibit similar behavior. Memory injections from conjunctions, however, typically outperformed all other parts of speech. We hypothesize that this is because conjunctions 
%a particularly common part of speech \cite{.}, 
often play a neutral role in prompt completions. Thus, while a random noun (e.g., \enquote{apple}) might distort prompt completion, a random conjunction (e.g., \enquote{and,} \enquote{for}) is less likely to do so.

We note also that for each part of speech, performance averaged over all injections for most ($\ell$,~$\tau$) pairs was reduced (< 0) for \easydata{} (refer Fig.~\ref{fig:pos_result}: subplots $c,d,g,h$), but was sometimes improved (> 0) for \harddata{} (refer Fig.~\ref{fig:pos_result}: subplots $a,b,e,f$). 
We attribute this result to the relative difficulties of the two datasets. 
\easydata{} has, on average, lower surprisals than does \harddata{}, as seen in Table~\ref{tab:data}, suggesting that there is additional information that the model could use successfully for \harddata{}, but not for \easydata{}.

%We note also that the prompt injections for our \easydata{} dataset, on average, always led to a degradation in prompt completion, while for \harddata{} they sometimes led to an improvement. We hypothesize that this is because of the relative difficulties of both datasets. 
%Our \easydata{} dataset has, on average, lower surprisals than does \harddata{}, as seen in Table~\ref{tab:data}. Thus, we hypothesize that while there is additional information that the model could use successfully for \harddata{}, this was not the case for \easydata{}.

These results (see also %more detailed heat maps are found in 
the Appendix; Figs~\ref{fig:gpt2_large_2wmh_pos}--% \ref{fig:gpt2_large_hand_pos}, \ref{fig:gpt2_small_2wmh_pos},
\ref{fig:gpt2_small_hand_pos}) suggest that while curated memories %at carefully selected 
%locations and magnitude 
are ideal for correcting multi-hop reasoning failures, language models can also benefit 
%to a degree from memories 
from injections of different parts of speech. %if injected in the correct manner. 
This result suggests 
%a division of labor in the language model: 
that different parts of a language model 
%(namely, different heads) 
(namely, early layers) serve specialized roles, 
%and some of those roles entail 
with some dealing with processing related to specific parts of speech.

In future work we will curate relevant memories from various parts of speech for each prompt, to better understand the effects of curated memories.

\section{Related Work}
\label{sec:related_work}
%From a high level
    % background on intermediate activations work
    % knowledge storage and editing in those activations
    % better predictions from edited activations
    % reasoning (multi-hop / chain of though)
    % alternative methods (knowledge graphs etc)

Much recent work has focused on the inner workings of Transformers \cite{vaswani2017attention,devlin2019bert,brown_LanguageModelsAre_2020,radford_llmsOpenAI_2019}.  \citet{nanda_ProgressMeasuresGrokking_2022} explore how the emergent properties of LLMs form during training. Recent interpretability research has focused on the mechanisms by which linear layers in LLMs retrieve information,  characterizing them as key-value stores of information \cite{geva_FeedForwardKeyValue_2021,dai-KnowledgeNeurons-2022,dai-NeuralKnowledgeBank-2022} and showing that tokens can be characterized by their distribution in the output vocabulary 
\cite{geva_FeedForwardBuildPredictions_2022}.

%We are not the first to 
Others have also examined the intermediate activations of LLMs in order to uncover 
%their 
underlying reasoning mechanisms. \citet{logitlens} 
%explored the mechanism by which GPT-2 arrives at its final answer in sentence completion tasks by applying the model's 
applied GPT-2's unembedding matrix to intermediate layers  to interpret how the model arrives at its final answer. \citet{tuned_lens} 
%further extended this method using 
employed a learned transformation to mitigate the effect of any bias introduced by using the unembedding matrix. 

There has been much recent interest in whether LLMs are reliable stores of information for attempting to both identify where knowledge exists and how to edit stored factual knowledge effectively \cite{mitchell_FastModelEditing_2022,mitchell_MemoryBasedModelEditing_2022,elazar-MeasuringAndImprovingConsistency-2021,hase-DoesLocalizationInformEditing-2023}. Recent approaches to knowledge editing make use of learned hyper-models to edit weights, additional trained parameters, or direct interventions on model weights \cite{decao-EditingFactualKnowledge-2021,huang-TransformerPatcher-2023,dhingra-TimeAwareLanguageModels-2022}. However, these approaches raise another issue: dealing with knowledge retention and preventing catastrophic forgetting \cite{jang-ContinualKnowledgeLearningLLMs-2022,hase-LanguageModelsHaveBeliefs-2021,mquake}. Additionally, it is not clear that the mechanisms by which model predictions are constructed is fully understood, limiting our ability to improve model performance \cite{turpin2023language}. Some approaches propose to use external knowledge stores such as knowledge graphs to augment the factual capabilities of LLMs \cite{jiang_unikgqa_2023,sun_open_2018,zhang_subgraph_2022}.

\section{Conclusions and Future Directions}
\label{sec:conclusions}
We demonstrate that a key reason LLMs perform worse on multi-hop prompts is because they fail to recall intermediary information that is relevant to a hop. We find that attention heads play an important role in this factual recall process, and that in the case of multi-hop reasoning, certain attention layers fail to recall relevant information.
To rectify this shortcoming, we establish an algorithm for injecting \enquote{memories} directly into the model's hidden activations during inference. Through experimentation, we find that injecting relevant memories into the hidden activations of the attention heads during inference is an efficient way to boost model performance on multi-hop prompts. 

We anticipate that our memory injection scheme can extend a model’s longevity by enabling less frequent retraining/fine-tuning. We also hope in future work to demonstrate the use of memory injections to correct stale or incorrect information, remove private or harmful information, and combat bias during LLM inference.

%Additionally, our memory injection scheme can extend a model's longevity as retraining/fine-tuning does not need to be done as often.

%We hope in future work to demonstrate how to use memory injections to correct stale or incorrect information, remove private or harmful information, and combat bias during LLM inference. 

There is also a tremendous opportunity to scale online-memory injections to enhance the quality of thousands/millions of model inferences, if we can automate the process of memory selection via unsupervised algorithms, for instance by connecting LLMs with knowledge bases. %One proposed approach to By co-locating models with external knowledge stores, such as knowledge graphs, we might be able to . [Need to provide a simple proposed frame work for how to automate memory selection, as it is the main bottleneck to being able to scale our work up.]

%This is beneficial because it is sometimes hard to know when a model is over/under parameterized so there is no guarantee that further fine-tuning was rectify the multi-hop failure in the same manner that our edits do. Furthermore, our edits are much more computationally efficient than further fine tuning. 

%Being able to apply edits to a model directly during weights also allows us to keep a model in use for longer; there is potential for future work to demonstrate how to efficiently correct stale or incorrect information using memory injections. 

% \section{Broader Impacts \& Ethics}
% \label{sec:ethics}
\section*{Limitations}
Internal biases of the question writers as well as the rigid structure that had to be imposed on the prompt structure mean that our human-generated dataset is representative only of a small fraction of the many types of multi-hop questions. Furthermore, our hand-generated dataset is relatively small compared to our programmatically generated dataset. Additionally, our analyses were limited to \gptsmall{} and \gptlarge{}; further work is needed to determine whether, as we expect, other language models sharing a transformer-based architecture and a similar unsupervised causal language modeling training objective display similar behavior. Lastly, we rely on the model's unembedding matrix $W_U$ to interpret model hidden states and embed \textit{memories} for injection. While for our work, results indicate that this transformation was sufficient, we acknowledge that this unembedding matrix is not tuned to interpret intermediate layers; we aim to address this shortcoming in future work by instead using layer-specific learned projections to transform between hidden states and vocabulary. 

\section*{Ethics}
Our attention head inspection mechanism uncovered several sources of bias (such as racism); refer Table~\ref{tab:atten_head_outputs} for examples. We expect a more detailed study of the attention heads of \gptsmall{} and \gptlarge{}, as well as other LLMs, to reveal additional undesirable behaviors. We aim in future work to use our inspection method to uncover (and hopefully address) these biases.

% \textbf{Security Concern:} The memory injection algorithm can be abused by a malicious party to inject backdoors at inference time. Furthermore, if the identities of the prompter are known, malicious injections could be tailored to attack specific targets.

%\textbf{Broader Impacts:} Memory injections can extend model longevity by allowing users to apply lightweight, non-gradient-based edits directly to the model's inference path. Thus, they can  reduce the need for costly model fine-tuning/re-training in order to meet standards for factual correctness or incorporate new information into an existing model. Additionally, memory injection can further augment the abilities of smaller LLMs, as smaller LLMs display a reduced capacity to store as much information as their larger counterparts. In this situation, memory injections, if applied correctly, may enhance the performance of AI in resource-constrained settings. As more robust and scalable methods for selecting memories are discovered in future work, memory injection can be adopted into existing inference workflows and as a means of augmenting LLMs with large knowledge stores.

\section*{Acknowledgements}
This material is based upon work supported by the U.S. Department of
Energy, Office of Science, Office of Advanced Scientific Computing Research, Department of
Energy Computational Science Graduate Fellowship under Award Number DE-SC0023112. 
This work is also supported in part by the U.S.\ Department of Energy under Contract DE-AC02-06CH11357.
% Acknowledgements will go here.

%TODO remove
% \nocite{*}
\bibliography{main}
\bibliographystyle{acl_natbib}

% \clearpage
% \newpage
%\newpage 
\onecolumn

\appendix

\section{Part-of-Speech Memory Injection Appendix}

\begin{figure*}[hbt!]
  \centering
      \includegraphics[width=\linewidth]{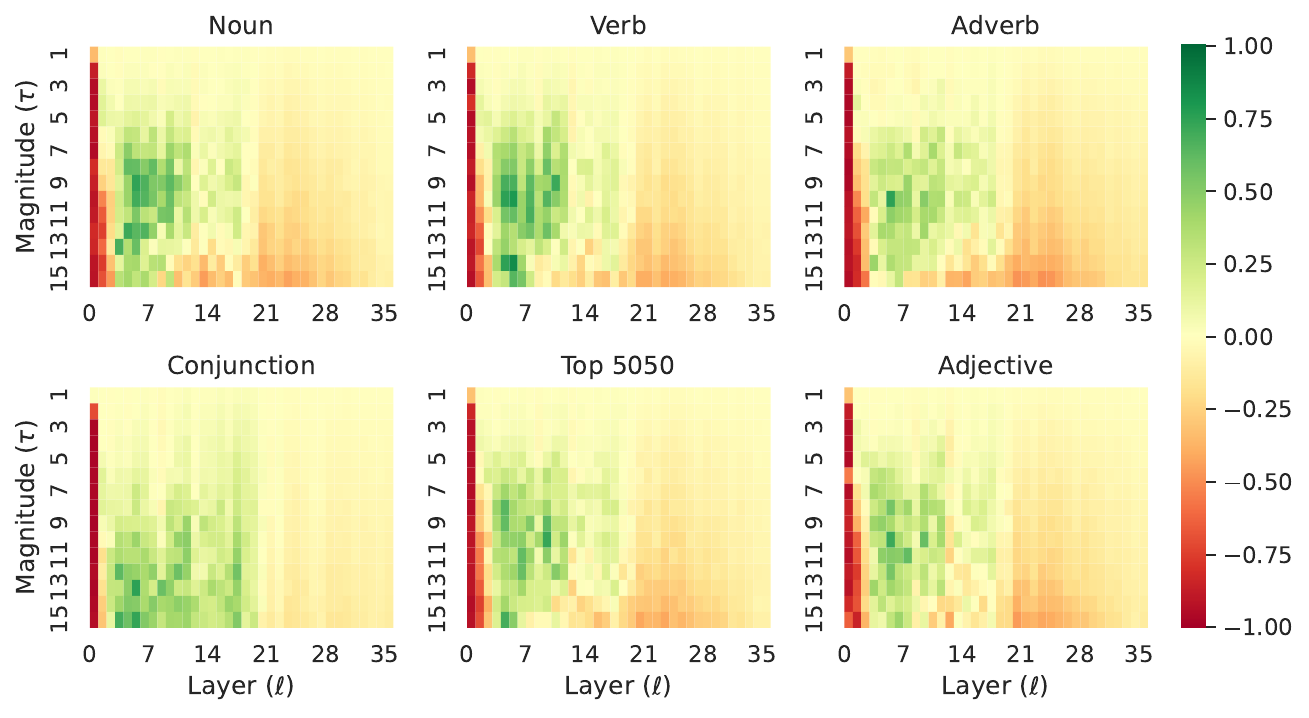}
  \caption{\textbf{GPT2-Large, \harddata{} dataset.} Heatmap shows average percent difference between pre- and post-injection answer probabilities for multi-hop prompts excluding outliers not within $\pm2$ standard deviations from the mean across various parts of speech.}
  \label{fig:gpt2_large_2wmh_pos}
\end{figure*}

\begin{figure*}[hbt!]
  \centering
      \includegraphics[width=\textwidth]{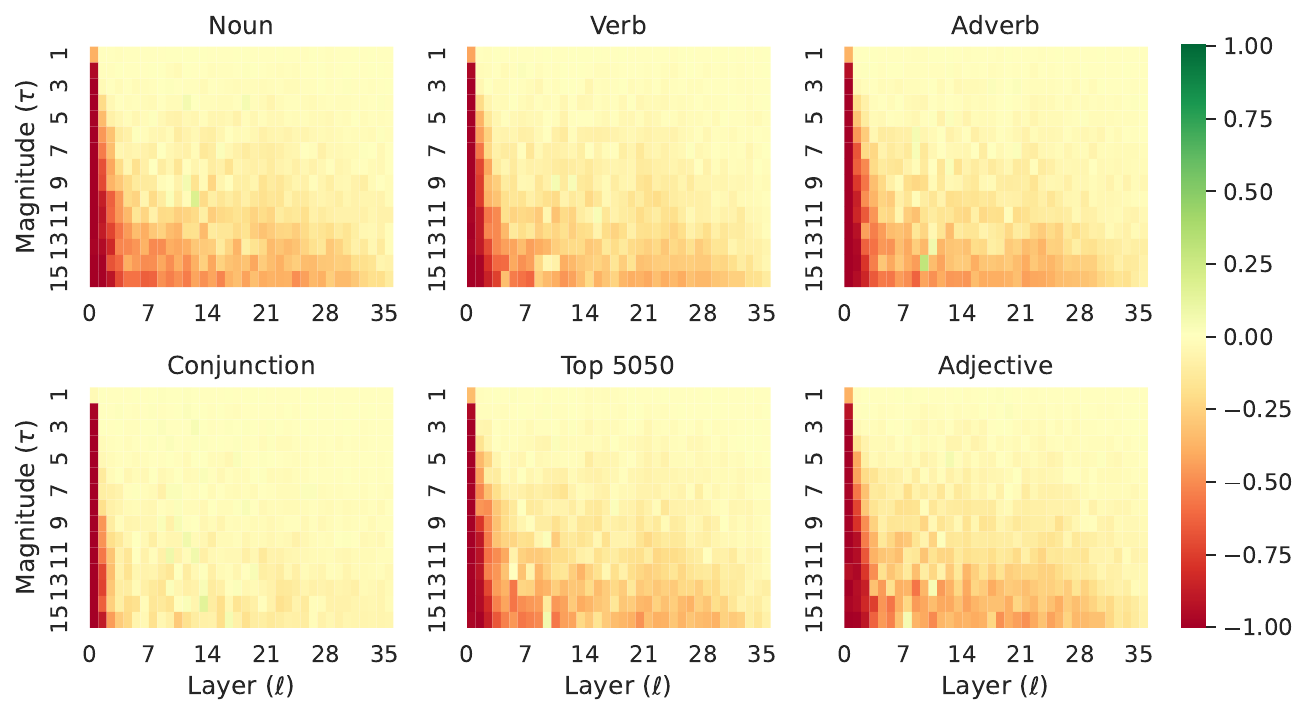}
  \caption{\textbf{GPT2-Large, \easydata{} dataset.} Heatmap shows average percent difference between pre- and post-injection answer probabilities for multi-hop prompts excluding outliers not within $\pm2$ standard deviations from the mean across various parts of speech.}
  \label{fig:gpt2_large_hand_pos}
\end{figure*}

%above $99^{th}$ percentile

\begin{figure*}[hbt!]
  \centering
      \includegraphics[width=\textwidth]{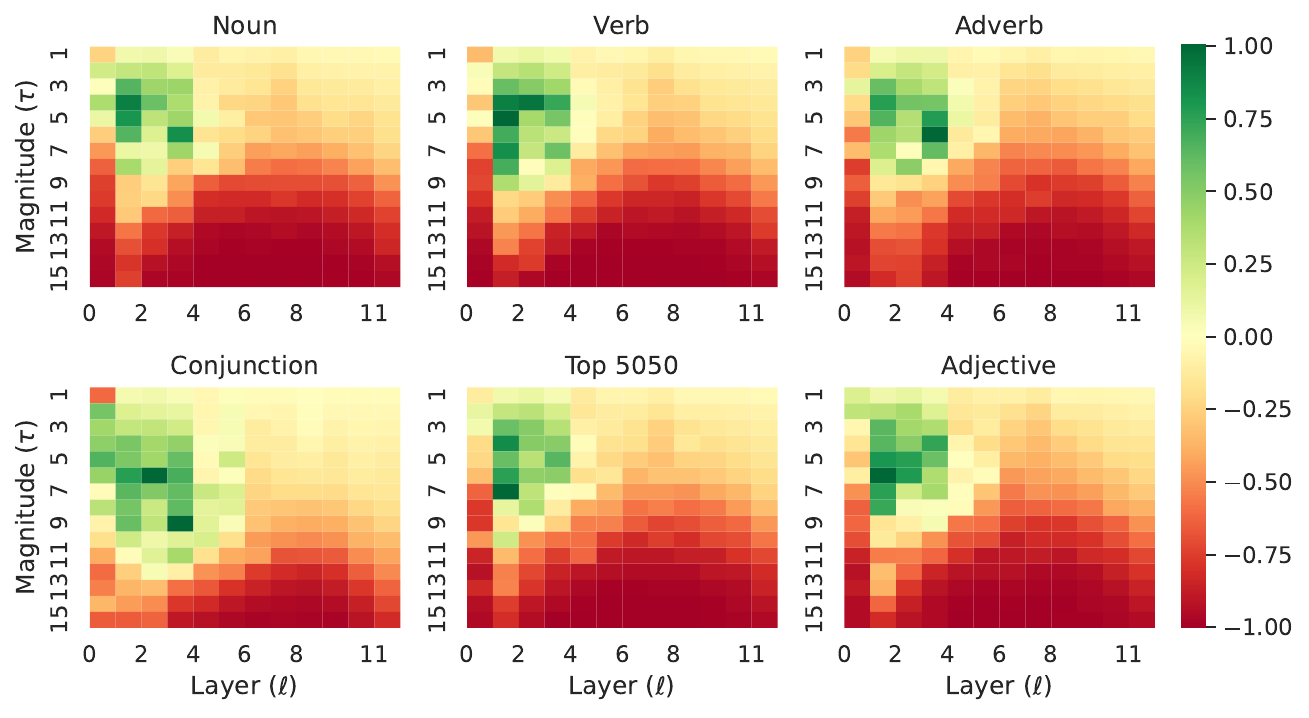}
  \caption{\textbf{GPT2-Small, \harddata{} dataset.} Heatmap shows average percent difference between pre- and post-injection answer probabilities for multi-hop prompts excluding outliers not within $\pm2$ standard deviations from the mean across various parts of speech.}
  \label{fig:gpt2_small_2wmh_pos}
\end{figure*}

\begin{figure*}[hbt!]
  \centering
      \includegraphics[width=\textwidth]{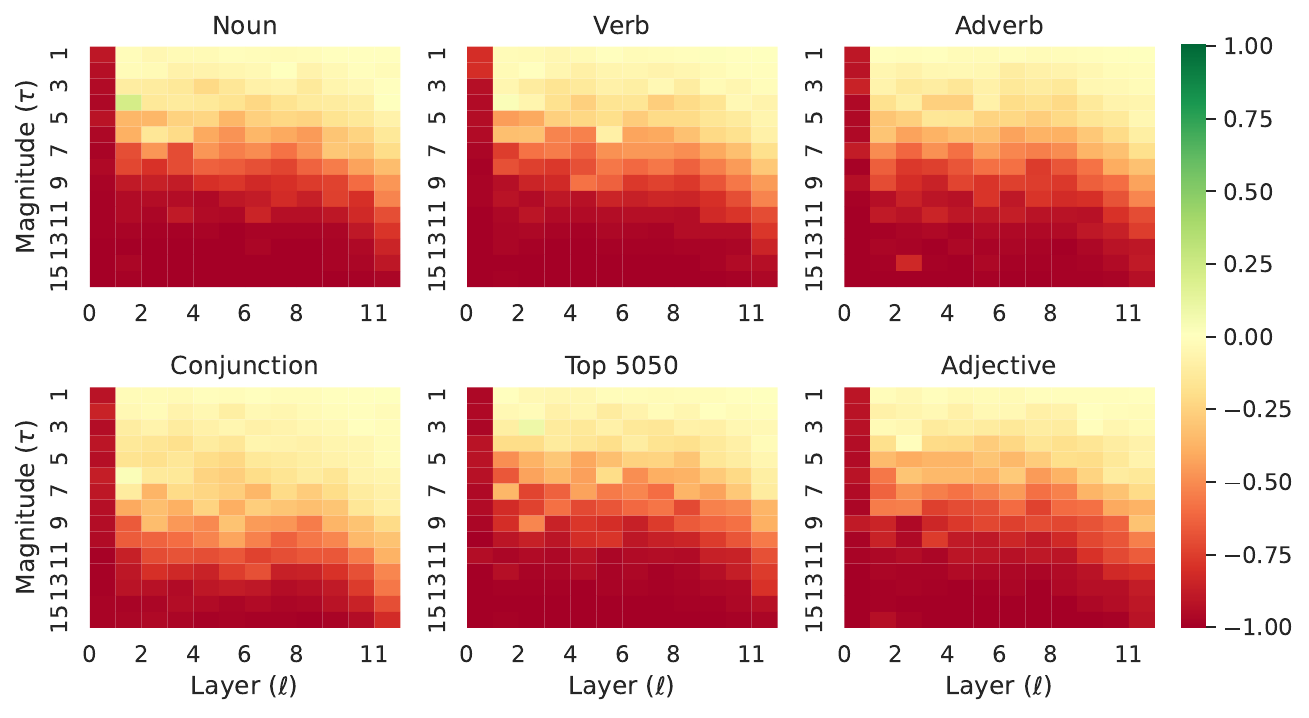}
  \caption{\textbf{GPT2-Small, \easydata{} dataset.} Heatmap shows average percent difference between pre- and post-injection answer probabilities for multi-hop prompts excluding outliers not within $\pm2$ standard deviations from the mean across various parts of speech.}
  \label{fig:gpt2_small_hand_pos}
\end{figure*}

\newpage

\null\newpage

\section{Dataset Example Appendix}
\begin{table*}[!htbp]
\centering
\small
\begin{tabular}{p{0.08\linewidth}p{0.4\linewidth}p{0.45\linewidth}}
\toprule
    Dataset & Single-Hop Prompt & Multi-Hop Prompt \\
\midrule
\multirow{9}{*}{\easydata{}}
    & George Washington fought in the \ldots [Revolutionary War] 
    & The first president of the United States fought in the \ldots [Revolutionary War] 
\\ \cmidrule{2-3}
    & Burj Khalifa is located in the city of \ldots [Dubai]
    & The tallest building in the world is located in the city of \ldots [Dubai]
\\ \cmidrule{2-3}
    & Nelson Mandela brought an end to \ldots [Apartheid]
    & The first president of South Africa brought an end to \ldots [Apartheid] 
\\ \cmidrule{2-3}
    & John F Kennedy was assassinated by a person named \ldots [Lee Harvey Oswald] 
    & The 35th president of the United States was assassinated by a person named \ldots [Lee Harvey Oswald]
\\ \cmidrule{2-3}
    & The father of Hermes is \ldots [Zeus] 
    & The father of the Greek messenger god is \ldots [Zeus] 
\\
\midrule
\multirow{10}{*}{\harddata{}}
    & The place of birth of Dušan Hanák is \ldots [Bratislava] 
    & The place of birth of the director of I Love, You Love is \ldots [Bratislava] 
\\ \cmidrule{2-3}
    & The employer of Éric Rohmer is \ldots [Cahiers du cinéma] 
    & The employer of the director of Triple Agent is \ldots [Cahiers du cinéma] 
\\ \cmidrule{2-3}
    & The employer of Chip Gubera is \ldots [University of Missouri] 
    & The employer of the director of Academy of Doom is \ldots [University of Missouri] 
\\ \cmidrule{2-3}
    & Steve Vai received the \ldots [Grammy] 
    & The performer of The Attitude Song received the \ldots [Grammy]
\\ \cmidrule{2-3}
    & The place of death of Augustus II the Strong is \ldots [Warsaw]
    & The place of death of the spouse of Christiane Eberhardine of Brandenburg-Bayreuth is \ldots [Warsaw]
\\
\bottomrule
\end{tabular}
\caption{\textbf{Example prompts.} Single/multi-hop prompt pairs from \easydata{} and \harddata{} datasets.}
\label{tab:prompt_examples}
\end{table*}
\begin{table*}[!htbp]
\small
\centering
\begin{tabular}{p{0.4\linewidth}p{0.15\linewidth}p{0.15\linewidth}p{0.1\linewidth}p{0.1\linewidth}}
\toprule
%     \multicolumn{4}{}{} & \multicolumn{7}{c}{Memory Injection}
% \\
% \cmidrule(lr){5-11}
%    \multicolumn{4}{}{} & Curated & \multicolumn{6}{c}{Random}
%\\

%\cmidrule(lr){5-5}
%\cmidrule(lr){6-11}
    Multiple-Hop Prompt & Memory & Answer & Pre-injection Answer Prob. &  Post-injection Answer Prob.
\\
\midrule 
    \textbf{The God of Thunder} is the son of \ldots & Thor & Odin & $0.84\%$ & $3.37\%$ 

\\
\midrule 
    \textbf{The first president to be assassinated} succeeded in abolishing \ldots & Abraham Lincoln & slavery & $30.46\%$ & $63.09\%$ 

\\
\midrule 
    \textbf{The founder of Microsoft} was born in the city of \ldots & Bill Gates & Seattle & $1.55\%$ & $2.44\%$  

\\
\midrule 
    \textbf{The highest peak in the world} is located in the \ldots & Mount Everest & Himalayan & $3.40\%$ & $22.58\%$ 

\\
\bottomrule
\end{tabular}
\caption{\textbf{Examples of memory injections.} Injecting memories with $\tau=4, \ell=9$ into \gptsmall{}.}
\label{tab:memory_injections}
\end{table*}
\begin{table*}[!htbp]
\centering
\small
\begin{tabular}{p{0.1\linewidth}p{0.2\linewidth}ccp{0.45\linewidth}}
\toprule
    Prompt Type & Prompt & Layer $\ell$ & Head $h$ & Output\\
\midrule
\multirow{29}{*}{Single-Hop}
    & John F Kennedy was assassinated by a person named \ldots
    & 10
    & 0
    & [` Kennedy', ` JFK', ` Assass', ` assass', `Kenn', ` assassination', ` Cuba', ` Oswald', ` assassin', ` Cuban', ` Fidel', ` Bobby', ` Havana', ` assassinated', ` assassins', ` Jackie', ` Castro', ` Jinn', ` assassinate', `Mu', ` 1963', ` Kahn', ` drone', ` Cah', ` Mu', ` Ghosts', ` Soul', ` Laos', ` Cemetery', ` CIA']
\\ \cmidrule{2-5}
    & Barack Obama was a member of the  \ldots
    & 9
    & 8
    & [` Obama', `Obama', ` Maryland', ` America', ` JFK', ` Biden', ` Harlem', ` Washington', ` American', ` Clinton', ` White', ` Americans', ` Congressional', ` Harvard', ` Kennedy', ` FBI', ` Federal', ` CDC', ` DOJ', ` President', ` Georgetown', ` HHS', ` Barack', ` US', ` Trayvon', ` Connecticut', ` Holder', ` New', ` BLM', ` Baltimore']
\\ \cmidrule{2-5}
    & Cain murdered a person named \ldots 
    & 2
    & 1
    & [` police', `,', ` the', ` a', `\textbackslash n', ` and', ` violence', `.', ` death', ` in', ` criminal', ` of', ` to', ` victim', ` "', `-', ` at', ` victims', ` crime', ` from', ` an', ` that', ` murder', ` crimes', ` is', ` was', ` he', ` for', ` (', ` killed']
\\ \cmidrule{2-5}
    & Russia is mostly located on the continent of \ldots 
    & 9
    & 8
    & [` Moscow', ` Russian', `Moscow', ` Russia', ` Kremlin', ` Putin', `Putin', `Russia', ` Russians', `Russian', `\encodingerror', ` \encodingerror', ` Dmitry', ` Mikhail', ` Vladimir', ` Sergei', ` Siberia', ` Soviet', ` Siberian', ` Ukraine', ` Ukrainian', ` Sochi', ` Caucasus', ` Nikol', `Soviet', ` KGB', ` Dmit', ` USSR', `Ukraine', ` Ukrainians']
\\ \cmidrule{2-5}
    & George Washington fought in the \ldots 
    & 9
    & 8
    & [` Washington', `Washington', ` Virginia', `Virginia', ` Maryland', ` Congressional', ` Georgetown', ` Dull', ` Smithsonian', ` Maine', ` Burr', ` Jefferson', ` Navy', ` Capitol', ` congressional', ` FDR', ` Lexington', ` Byrd', ` Rhode', ` Roosevelt', ` Pike', ` Everett', ` Brookings', ` Madison', `apeake', ` Randolph', ` VA', ` Arlington', ` Americans', ` Lafayette']
\\
\midrule
\multirow{26}{*}{Multi-Hop}
    & The 35th president of the United States was assassinated by a person named \ldots 
    & 10
    & 0
    & [` assass', ` Assass', ` assassination', ` assassin', ` assassins', ` assassinate', ` Malik', ` bullets', ` gunmen', ` assassinated', `Mu', ` Pakistani', ` sniper', ` killings', ` JFK', ` Pakistan', ` homicides', ` Alger', ` lethal', ` Islamabad', ` Karachi', ` shooting', ` gun', ` gunshot', ` Mu', ` murder', ` killing', ` pistols', ` murders', ` gunned']
\\ \cmidrule{2-5}
    & The first black president of the United States was a member of the  \ldots
    & 9
    & 8
    & [` Negro', ` NAACP', ` blacks', ` black', ` Baltimore', ` White', ` negro', ` Washington', ` BLM', ` white', ` FBI', ` America', ` Maryland', ` African', ` Trump', ` Nixon', ` Charleston', ` Americ', ` KKK', `Washington', ` Virginia', ` racial', ` Blacks', `white', `White', ` nig', ` Black', ` Obama', ` Louisiana', ` whites']
\\ \cmidrule{2-5}
    & Adam and Eve's eldest son murdered a person named \ldots 
    & 2
    & 1
    & [`,', ` the', ` and', ` a', ` "', ` in', `\textbackslash n', `.', ` to', ` of', ` at', ` is', ` he', `-', ` that', ` was', ` for', ` police', ` from', ` on', " `", ` as', ` death', ` had', "'", ` an', ` his', "'s", ` said', ` told']
\\ \cmidrule{2-5}
    & The largest country in the world is mostly located on the continent of \ldots 
    & 9
    & 8
    & [`,', `\textbackslash n', ` the', ` and', `.', ` in', ` a', ` to', ` of', ` (', `-', ` for', ` that', ` "', `:', ` is', ` or', ` at', ` as', ` I', ` on', ` with', ` it', ` an', ` from', ` all', ` by', ` not', "'s", ` more']
\\ \cmidrule{2-5}
    & The first president of the United States fought in the \ldots 
    & 9
    & 8
    & [` Trump', ` Washington', ` America', `Washington', ` American', `Trump', `America', ` Obama', ` Donald', ` FBI', ` Congressional', ` Americans', `American', ` Nixon', ` Congress', ` congressional', ` White', ` Roosevelt', ` Republican', ` Negro', ` Clinton', ` JFK', ` Reagan', ` Virginia', ` FDR', `Obama', `Americans', ` Americ', `FBI', `Congress']
\\
\bottomrule
\end{tabular}
\caption{\textbf{Example of attention head outputs} from \gptsmall{} for \easydata{}.}
\label{tab:atten_head_outputs}
\end{table*}

\end{document}